\documentclass[HARVARD,Times1COL]{WileyNJDv5} 
\usepackage[caption=false]{subfig}
\usepackage{graphicx}
\usepackage{makecell}

\articletype{Article Type}%

\received{Date Month Year}
\revised{Date Month Year}
\accepted{Date Month Year}
\journal{Journal}
\volume{00}
\copyyear{2026}
\startpage{1}

\begin{document}

\title{Multiple Object Detection and Tracking in Panoramic Videos for Cycling Safety Analysis}

\author[1]{Jingwei Guo*}

\author[2]{Yitai Cheng*}

\author[2]{Meihui Wang}

\author[2]{Ilya Ilyankou}

\author[2]{Natchapon Jongwiriyanurak}

\author[2,3]{Xiaowei Gao}

\author[4]{Nicola Christie}

\author[2]{James Haworth}

\authormark{JINGWEI \textsc{et al.}}
\titlemark{Multiple Object Detection and Tracking in Panoramic Videos for Cycling Safety Analysis}

\address[1]{\orgdiv{Department of Civil, Environmental, and Geomatic Engineering}, \orgname{University College London}, \orgaddress{\state{Gower Street, London}, \country{United Kindom}}}

\address[2]{\orgdiv{SpaceTimeLab, Department of Civil, Environmental, and Geomatic Engineering}, \orgname{University College London}, \orgaddress{\state{Gower Street, London}, \country{United Kindom}}}

\address[3]{\orgdiv{Department of Earth Science and Engineering}, \orgname{Imperial College London}, \orgaddress{\state{Exhibition Road, London}, \country{United Kindom}}}

\address[4]{\orgdiv{Centre for Transport Studies, Department of Civil, Environmental, and Geomatic Engineering}, \orgname{University College London}, \orgaddress{\state{Gower Street, London}, \country{United Kindom}}}

\corres{James Haworth
\email{j.haworth@ucl.ac.uk}}

\fundingInfo{This research was funded by the Road Safety Trust (RST 38\_3\_2017) and supported by a UCL Chadwick Scholarship.}

\abstract[Abstract]{Cyclists face a disproportionate risk of injury, yet conventional crash records are too sparse to identify risk factors at fine spatial and temporal scales. Recently, naturalistic studies have used video data to capture the complex behavioural and infrastructural risk factors. A promising format is panoramic video, which can record 360$^\circ$ views around a rider. However, its use is limited by distortions, large numbers of small objects and boundary continuity, which cannot be handled using existing computer vision models. This research proposes a novel three-step framework: (1) enhancing object detection accuracy on panoramic imagery by segmenting and projecting the original 360$^\circ$ images into sub-images; (2) modifying multi-object tracking models to incorporate boundary continuity and object category information; and (3) validating through a real-world application of vehicle overtaking detection. The methodology is evaluated using panoramic videos recorded by cyclists on London’s roadways under diverse conditions. Experimental results demonstrate improvements over baselines, achieving higher average precision across varying image resolutions. Moreover, the enhanced tracking approach yields a 10.0\% decrease in identification-switches and a 2.7\% improvement in identification precision. The overtaking detection task achieves a high F-score of 0.82, illustrating the practical effectiveness of the proposed method in real-world cycling safety scenarios.}


\maketitle

\renewcommand\thefootnote{}
\footnotetext{The email address of the submitting author: yitai.cheng.21@ucl.ac.uk.}
\footnotetext{*These authors contributed equally to this work.}

\renewcommand\thefootnote{\fnsymbol{footnote}}
\setcounter{footnote}{1}

\section{Introduction}

Cycling, as an active travel mode, has zero carbon emissions \citep{massink_2011} and can support reductions in traffic congestion \citep{brunsing_1997} while improving the health of urban residents \citep{wanner2012active}. Therefore, many local governments and transport authorities are implementing policies to encourage cycling as an alternative to car travel. However, cycling is often perceived as a dangerous activity, particularly in cities that lack high-quality cycling infrastructure. In London, cycle lanes account for only 6.4\% of the total road length (excluding highways) \citep{tait2022cycling}, meaning that cyclists often have to share lanes with motor vehicles.  According to \cite{transport_for_london_casualties_2020}, despite the mode share of cyclists being only 3.4\%, there were 868 killed or seriously injured (KSI) incidents involving cyclists, accounting for 28\% of all such casualties in the city in 2020. This disproportionately high risk of incidents can reduce people's willingness to ride \citep{Delmelle2012}, which hinders the uptake of cycling. Thus, it is necessary to take measures to improve urban cycling safety, which may include improving road infrastructure \citep{pucher2016safer}, implementing policies such as lower speed limits, or running public awareness campaigns like Operation Close Pass\footnote{https://www.cyclinguk.org/article/police-close-pass-day-action-tackle-dangerous-overtaking}. However, to gain support, these strategies and improvements must be accompanied by evidence of their effectiveness. One aspect of this is the impact on cycling safety.

Although crash data are available through hospital reports and official statistical releases, such as the UK’s STATS19, the low mode share of cycling makes it difficult to draw statistically significant conclusions at a spatially and temporally granular level. Near-miss events, in contrast, are more frequent and have provided a rich source of data for researchers in the form of naturalistic studies \citep{Ibrahim2021}. These studies have used a range of sensors to collect data while people carry out routine activities on bikes, with the rider indicating when they experience a near-miss. Alternatively, sensors can collect objective data on a near-miss type, such as a close pass \citep{BECK2019253}. In recent years, with the rapid development of street view imagery, action/dash cameras, virtual reality (VR) and related technologies, panoramic photography is becoming widespread and mature \citep{Im2016}. Unlike traditional cameras, which can only capture content within a limited field of view, a panoramic camera can capture 180-degree views of the front and rear of the device at the same time with two fisheye lenses, and then concatenate them into one high-resolution 360$^\circ$ image \citep{hong_hua_2001}. These consumer-grade cameras are suitable to be mounted on helmets or handlebars to record videos of the riders’ surroundings in all directions. Importantly, the 360 view captures the relative movement of the rider and the other people and objects recorded in the scene. For example, in an overtaking manoeuvre, a vehicle will first appear in the rear-view of the camera, before passing into the side view (right or left depending on locality) and the front view. Computer vision based object detection and tracking technologies have the potential to track these movements and categorise them into actions based on interpretations of the relative movements.   

The emergence of deep learning has had a transformative effect on a range of machine learning tasks, especially computer vision \citep{LeCun2015}. A range of algorithms have been developed for implementing various visual tasks, such as object detection \citep{zhao2019object} and multiple object tracking (MOT) \citep{Luo2021}. However, since most of these models and the attached pre-trained weights were designed and trained for traditional images and videos, if applied directly to panoramic videos in equirectangular projection, their performance will be negatively affected \citep{yang2018object}. In addition, some models' evaluation metrics do not apply to panoramic images. These problems exist because of the following differences between panoramic images and standard images: 

\begin{itemize}
\setlength\itemsep{0pt} 
\setlength\parskip{0pt}  
    \item The acquisition of panoramic videos requires specific cameras. As a result, the number of available annotated panoramic datasets is much smaller than that of standard cameras;
    \item Due to the distortions introduced by the equirectangular projection, objects near the camera may be seriously deformed, while objects far away from the camera will be small; 
    \item Panoramic images in equirectangular projection have boundary continuity in which the contents of the left and right sides of the image are continuous \citep{liu_2018}. 
\end{itemize}

To overcome these limitations, this research develops a framework for applying deep learning based computer vision models pre-trained on `traditional' datasets to the task of object detection and tracking in panoramic videos. The methodology is applied to a case study in cycling safety analysis; the identification of overtaking manoeuvres.

The remainder of the paper is structured as follows: Following this Introduction, we discuss related work on cycling safety, computer vision and its application to panoramic data in Section \ref{RelatedWork}. Section \ref{Data} describes the collection, pre-processing and annotation process of the data used in this project, and introduces their distribution. In Section \ref{Methodology}, we introduce the methodology, and then present and analyse the results in Section \ref{Results}. In Section \ref{Discussion}, we discuss the successes and limitations of our work and propose some future work to overcome the limitations. Finally, we present our conclusion in Section \ref{Conclusion}.

\section{Related work} \label{RelatedWork}

Urban transport networks are complex environments in which many agents (pedestrians, cyclists and other road users) interact with each other while using different types of infrastructure. These actions and interactions within the environment contribute to crash risk. Traditionally, crash risk is assessed using historical data on crash records, such as the UK's STATS19, combined with measures of exposure such as traffic flows, which are combined in a multiple regression framework to identify risk factors \citep{Ambros2018,gao2025reliable}. Existing cycling safety studies have heavily relied on crash and conflict data from open government resources \citep{Hels2007, Daniels2008, zhang2023analyzing}, questionnaires surveys \citep{wang2020exploring, useche2022cross}, and crowdsourced data reported by volunteer cyclists \citep{poulos2012exposure, fischer2020does}. However, these data sources are often subjective and incomplete, making understanding the underlying causes of hazards challenging \citep{gao2025reliable}. Furthermore, due to the low mode share of cycling in many countries, records on bicycle crashes may be sparse, and exposure data may be limited to particular locations and times or not collected at all. 

To complement traditional methods, researchers have turned to naturalistic studies to observe cyclists as they carry out their routine activities \citep{Ibrahim2021}. Typically, this involves the rider using a suite of sensors that passively collect data such as video, inertial measurement unit (IMU) and location data. Early work relied on customised equipment for this purpose \citep{johnson_2010}, which limited its applicability beyond the scope of the study. Subsequent studies produced sensors for particular purposes, such as detecting close passes, enabling the collection of high-quality, objective data \citep{BECK2019253}. \cite{Ibrahim2021} provide a review of the recent work. While customised equipment allows a high level of control over experimental design, the use of consumer-grade equipment provides an opportunity for larger scale data collection. For example, many road users routinely use dashcams or action cameras to collect data on their journeys to protect themselves if they are involved in a crash. Recognising this, recent work, particularly in autonomous vehicles, has leveraged dashcam video data for crash detection. \cite{rocky_review_2024} provides a review of the current state of the art. Similar methods have also been applied in cycling safety analysis. For example, using data sourced from YouTube, \cite{ibrahim_cyclingnet_2021} used a deep learning approach that harnessed the optical flow of video data to detect near misses from video streams. Harnessing this data as a form of crowdsourcing could provide a way to monitor safety from the cyclist's perspective continuously, and studies such as \cite{rick_cycling_2021} have used popular devices such as GoPro cameras to this end.

The key enabling computer vision technologies for understanding the risk environment from video streams are object detection and multiple object tracking (MOT). Object detection identifies and localises objects, such as vehicles and people, within a video frame. Techniques such as You Only Look Once (YOLO) and Faster Residual Convolutional Neural Network (R-CNN) have been employed that achieve high accuracy in detecting different road users, such as CyDet \citep{masalov_2018} and MoDeCla \citep{ojala2022motion}. MOT is a task that associates multiple objects in different frames of a video into several trajectories \citep{JimnezBravo2022}. MOT technologies have been widely used in the field of transport, particularly in autonomous vehicles. Various modifications have been made to improve car tracking in poor light conditions \citep{Taha2015}, continuously track road users that are sometimes occluded in the videos \citep{jodoin_2014}, and to track objects using the views from multiple cameras \citep{JimnezBravo2022}.

Existing object detection and MOT algorithms are trained on video from cameras with a limited field of view (FOV). This suits the data that has been collected in naturalistic cycling studies to date, which contain only front or rear-facing views. However, risks that cannot be captured in a single view may be present. For example, rear-end crashes have proven to result in a high fatality rate, and tailgating is a risk factor \citep{australian_transport_safety_bureau_deaths_2006}. For a comprehensive view of the risk environment, recording video both in front of and behind the rider is necessary. This can be accomplished using separate cameras, but leads to problems of synchronisation and matching of objects between the two video streams. Furthermore, the side views of the rider are not captured, which are important in the context of certain events such as close passes and car-dooring. Alternatively, panoramic cameras capture 360$^\circ$ views around the user, allowing seamless tracking of interactions between the rider, the environment and other road users. 

Existing CV models struggle with panoramic videos due to a lack of panoramic training data, distorted and small objects in equirectangular projections, and issues with boundary continuity. Various studies have attempted to solve these issues. Some have created panoramic datasets to train existing models \citep{deng_2017, zhang2017real}, while others have applied techniques to improve model performance without retraining \citep{yang2018object}. Still, these do not fully address boundary continuity or the challenge of splitting long objects across subwindows. For MOT, the quality of object detection significantly affects tracking performance. StrongSORT is a commonly used tracker that has been adapted for panoramic data. However, it faces challenges such as object ID switching caused by boundary continuity and inaccuracies in multi-category tracking (MCT). \citep{liu_2018} proposed extending input images to mitigate boundary continuity issues, yet further improvements are needed to handle objects appearing multiple times across frame boundaries and to enhance MCT.

To address these issues, this research combines and extends these approaches by adding support for boundary continuity and long objects in detection, introducing category information to reduce tracking errors, and enhancing boundary continuity implementation in StrongSORT for better multi-object tracking.


\section{Data} \label{Data}
To test the effect of the proposed methods on object detection and MOT on panoramic cycling videos, this project built an annotated dataset. In addition, some other video clips containing overtaking were prepared to present and evaluate the application results of the improved models in recognising the overtaking of the surrounding vehicles.

\subsection{Data collection}
To evaluate the method for multiple object detection and tracking, panoramic videos were collected by the research team while cycling in London, UK. A GoPro Max camera was attached to the rider's helmet using a metal mount. No restrictions were placed on when and where to ride --- the researcher recorded videos while carrying out routine activities such as commuting. The only limitation was that the data collection had to be in good lighting conditions so that all the objects would be recognisable in the videos. This is almost always the case in urban settings due to street lighting. Since the dataset with annotations was built only for evaluating the proposed models rather than training from scratch, there was no requirement for a large data volume. Therefore, 3 clips were chosen from the collected videos, which are 10 to 15 seconds in length.


To mitigate bias that may be introduced by using video from a single person in a limited range of conditions, validation for overtaking detection is carried out on a set of video clips filmed by 49 different cyclists who were participants in UCL's 100 Cyclists Project \citep{christiedetecting}. 
This dataset provides a broader range of riding conditions and rider characteristics on which to test the proposed method: 50 of the videos contain a single clear overtaking event lasting several seconds (`positive' examples), and 50 do not (`negative'). To account for diverse cycling environments, selected videos were captured during daytime (78), night-time (17), and at dawn or dusk (5). 24 videos featured glare from the sun and artificial lighting sources, such as lampposts and headlights. While only one video was recorded in the rain, a further 6 were recorded after recent rain and are slightly obfuscated by raindrops on the camera lens.


\subsection{Data Pre-processing \& Annotation} \label{DataPreprocessing}


Videos were trimmed and formatted to `Horizon Levelling' by using gyroscope data to level the videos on the horizontal axis. After that, the clips were exported as 5.6k (5,368 $\times$ 2,688) equirectangular videos.

To generate labels to evaluate object detection and MOT, the 3 video clips collected by the research team were annotated using the Computer Vision Annotation Tool (CVAT)\footnote{https://github.com/cvat-ai/cvat}. The following 7 categories of objects were chosen for labelling, which are both relevant to cycling safety and contained in the Microsoft COCO dataset \citep{lin_2014}: person, bicycle, car, motorbike, bus, truck and traffic light. In order to output the annotations in both COCO and MOT formats, `tracks' were created to associate the same objects in different image frames.

Following the standards of MOT format, CVAT only supports labelling each object once in each frame. Hence, objects are divided into two parts due to the boundary continuity could be labelled as two separate objects. To address this, after the annotations were output in both COCO and MOT formats, a method was implemented to merge these split objects in the MOT annotation files without violating the rules of MOT format \citep{milan2016mot16benchmarkmultiobjecttracking}. After the ID modifications, videos were split into image frames. These images are used to evaluate object detection models with the COCO annotation files. Finally, after pre-processing and annotation, the dataset has 24,454 labelled objects and 224 labelled tracks in total. 

\subsection{Exploratory Data Analysis} \label{eda}

In order to design the methodologies better for improving the applicability of the existing object detection and MOT models and to better understand the corresponding evaluation results, this project conducted some exploratory analysis of the annotated dataset. 

\begin{figure}
    \centering
    \includegraphics[width=0.6\linewidth]{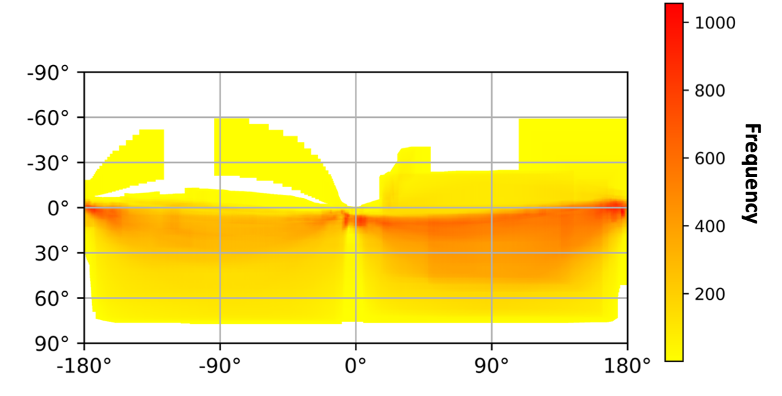}
    \caption{Position distribution of the ground truth bounding boxes in the dataset}
    \label{fig:fig2}
\end{figure}

To understand where the objects of interest are located, a matrix whose size is the same as the videos (5,368 $\times$ 2,688) is used to count how many times each pixel is within an annotated bounding box. Then, the matrix is plotted as a heat map, as Figure \ref{fig:fig2} shows. As the top and bottom areas of an equirectangular image represent the sky and the ground, respectively, almost all the objects in the dataset are distributed from -50$^\circ$ to 70$^\circ$ of the y axes.

\section{Methodology} \label{Methodology}
The proposed methodology has three steps. In step one, a method is developed to enhance the applicability of existing pre-trained object detection models to equirectangular images. In step two, these models are set as the detector of StrongSORT, and its structure is modified according to the characteristics of panoramic videos. Finally, in step three, a downstream application is developed to automatically detect overtaking behaviour in panoramic videos.  

\subsection{Improving Object Detection on Panoramic Images} \label{ImprovingObjectDetection}

\begin{figure}[!t]
    \centering
    \includegraphics[width=0.9\textwidth]{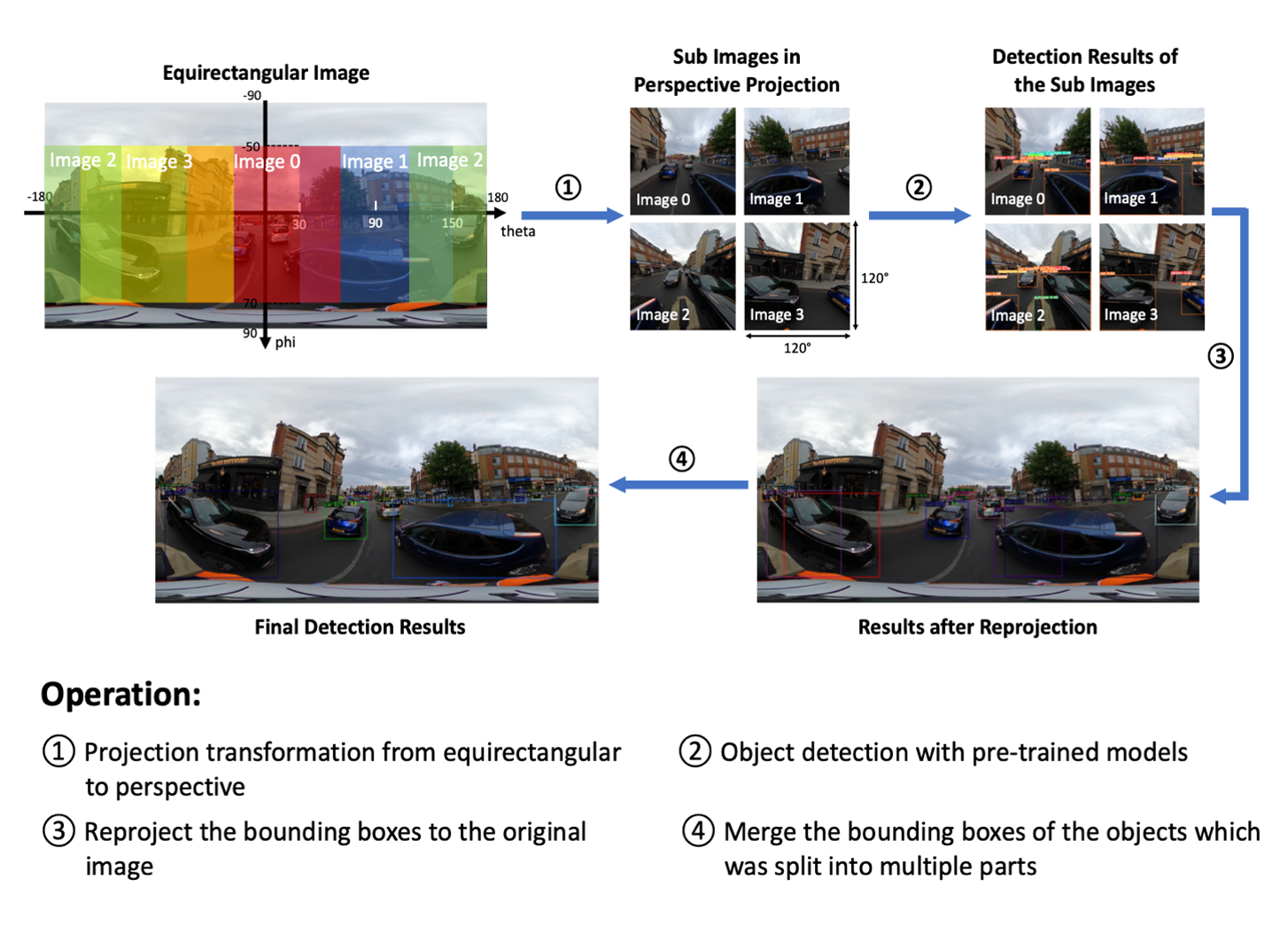}
    \caption{Workflow of the proposed method for improving object detection on panoramic images.}
    \label{fig:fig3}
\end{figure}

The proposed method for improving object detection on panoramic images, shown in Figure \ref{fig:fig3}, consists of the following four operations: 1) Projection of equirectangular images into four sub-images of perspective projection; 2) implementation of object detection models pre-trained on the COCO dataset on each of the sub-images; 3) Reprojection of the detected bounding boxes in the sub-images to the original image; 4) Merging bounding boxes of  long objects (such as the blue and red cars in Figure \ref{fig:fig3}). 

In the following subsections, each of these steps are explained in detail. After that, the evaluation method is introduced.  

\subsubsection{Projection Transformation from Equirectangular to Perspective}

To present a whole sphere as a 2D image, equirectangular projection maps longitude lines and latitude circles to equally spaced vertical and horizontal lines, respectively \citep{snyder1997flattening}. In contrast, perspective projection only maps part of a sphere (within the FOV) from the centre to a tangent plane \citep{yang2018object}. Thus, the transformation from an equirectangular to a perspective is a projection from a sphere to a plane. 

According to \cite{Lee2021}, assuming that the perspective projection is along the z-axis, the corresponding point P' $(X, Y, Z)$ on the sphere of the point P $(u, v)$ on the tangent plane can be calculated as:

\begin{equation}
X = \frac{u - T}{\sqrt{u^2 + v^2 + 2T^2 -2uT - 2vT + 1}}
\label{eq8}
\end{equation}

\begin{equation}
Y = \frac{-v + T}{\sqrt{u^2 + v^2 + 2T^2 -2uT - 2vT + 1}}
\label{eq9}
\end{equation}

\begin{equation}
Z = \frac{ 1 }{\sqrt{u^2 + v^2 + 2T^2 -2uT - 2vT + 1}}
\label{eq10}
\end{equation}

Where $T$ is the tangent of the half of the field of view of the projection, i.e., $T = tan(\frac{FOV}{2})$.

If represented in a geographic coordinate system, P' can be also represented as longitude $\phi$ and latitude $\theta$ \citep{Lee2021}: 

\begin{equation}
    \phi = tan^{-1}(-\frac{Z}{X})
    \label{eq11}
\end{equation}

\begin{equation}
    \theta = sin^{-1}(\frac{Y}{\sqrt{X^2 + Y^2 + Z^2}})
    \label{eq12}
\end{equation}

If the projection is not towards the z-axis, when calculating the coordinates of point P' $(X', Y', Z')$, a rotation equation $R$ should be applied as follows \citep{Lee2021}:

\begin{equation}
    \begin{bmatrix}
        X'\\ Y'\\ Z'\\
    \end{bmatrix}
    =  R \begin{bmatrix} X\\ Y\\ Z\\
    \end{bmatrix}
    =
    \begin{bmatrix}
    cos\alpha & -sin \alpha sin\theta_c & sin\alpha cos \theta_c \\
    0 & cos\theta_c & sin\theta_c \\
    -sin\alpha & -sin\theta_c cos\alpha & cos \alpha cos \theta_c
    \end{bmatrix}
    \begin{bmatrix}
    X \\ Y \\ Z
    \end{bmatrix}
    \label{eq13}
\end{equation}
where $\alpha = \phi_c + \frac{\pi}{2}$, $\phi_c$ is the angle between the projection direction and plane $xoz$, while $\theta_c$ is the angle between the projection direction and plane $yoz$. 

The transform from spherical coordinate to Equirectangular representation is as follows: 
\begin{equation}
    m = \frac{\phi + \pi}{2\pi}W, \
n = \frac{\tfrac{\pi}{2} - \theta}{\pi}H
    \label{eq14}
\end{equation}
where \((m,n)\) represents a pixel in ERP format. \(m\) denotes its horizontal position and \(n\) denotes its vertical position. \(W\) represents the width of the image in ERP format and \(H\) represents the height. 

For each output pixel in the perspective projected image, its corresponding location in the original ERP format is computed by Equations (\ref{eq8}) to (\ref{eq14}). Then the value of the pixel is computed by cubic interpolation where a 4 by 4 block of its neighbouring pixels is involved.
To identify parameters for perspective projection transformation, this research referred to the position distribution analysed in subsection \ref{eda}: since almost all the objects in the videos are distributed from -50$^\circ$ to 70$^\circ$ vertically, FOV and $\phi_c$
were set to 120$^\circ$ and -10$^\circ$, so that only the objects within this interval would be projected and detected in the improved object detection model. 

Although three sub-images with a 120$^\circ$ FOV are enough to cover an equirectangular image horizontally, because there are still some distortions in the perspective images, objects near the edges of the sub-images may not be detected. To solve this problem, the original image is projected into 4 sub-windows along the lines of 0$^\circ$, 90$^\circ$, -180$^\circ$ and -90$^\circ$ longitude (i.e., theta) --- since there is an overlap of 30$^\circ$ between each pair of adjacent sub-images, objects within 15$^\circ$ from an edge in one sub-image are closer to the centre in another sub-image, which results in less severe distortions.

\begin{figure}[!t]
\centering

\subfloat[Process of reprojecting a bounding box from perspective to equirectangular\label{fig:fig4}]{
    \includegraphics[width=0.49\textwidth]{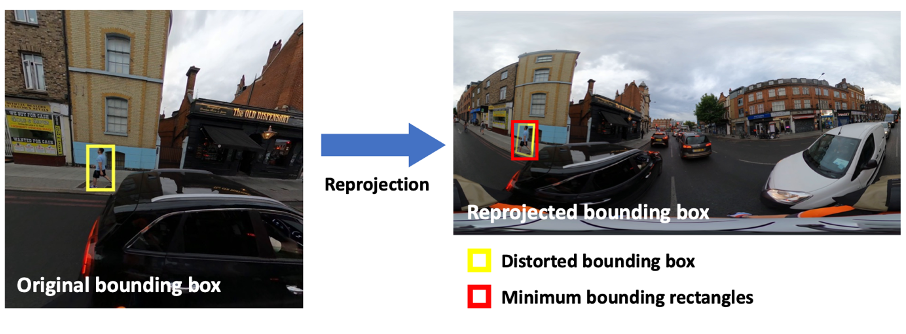}
}
\hfill
\subfloat[Process of reprojecting an object which is across the boundaries\label{fig:fig5}]{
    \includegraphics[width=0.49\textwidth]{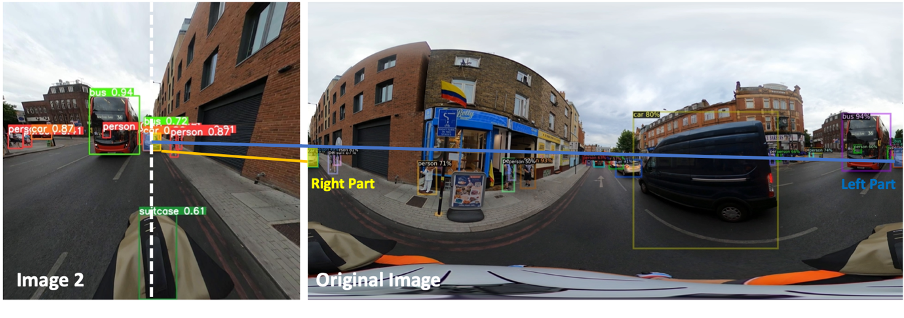}
}

\vspace{1em} 

\subfloat[Method for processing the objects which are displayed in several sub-images\label{fig:fig6}]{
    \includegraphics[width=0.8\textwidth]{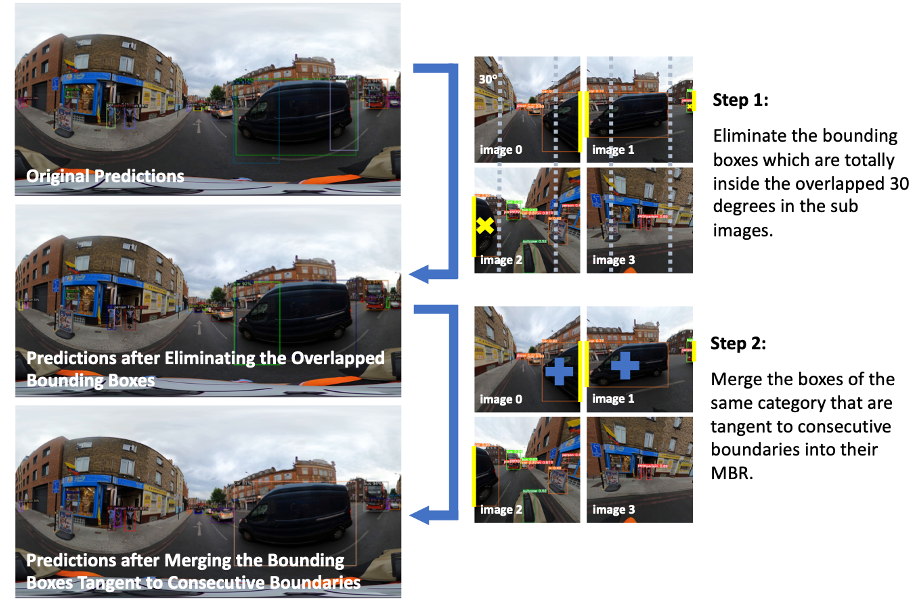}
}

\caption{Projection and aggregation of the detection results of sub-images.}
\label{fig:three-images}
\end{figure}

\subsubsection{Object Detection with Pre-trained Models}

For the object detection model, this study uses YOLO12\citep{tian2025yolov12}, which has state-of-the-art object detection accuracy at the time of writing. In this step, YOLO12n pre-trained on the COCO dataset was chosen since it is light-weight but still performs well on the validation dataset. 

\subsubsection{Reprojection of the Detection}

When projecting a perspective sub-image back onto an equirectangular panorama, we precompute two 2D lookup tables --- one encoding the horizontal (longitude) index and the other the vertical (latitude) index in the equirectangular frame for every pixel in the perspective view.  Then, for any pixel at coordinates (u,v) in the perspective image, its corresponding sample location (m,n) in the panorama is simply given by the values stored in those two tables at position (u,v). This general approach disentangles the projection math from the sampling step, allowing the perspective view to be assembled by a straightforward indexed lookup into the equirectangular source.

As Figure \ref{fig:fig4} shows, to map all the bounding boxes, which are at least 15$^\circ$ away from the edges back to the original image, every point on the borders of these boxes was reprojected with the approach above. However, this process distorts the edges of the boxes, which makes it difficult to output and evaluate the predictions. Therefore,  minimum bounding rectangles are used represent the boxes. 

Due to the boundary continuity issue, detections crossing the centre lines of the sub-images projected along 180$^\circ$ are shown twice near the left and the right edges, respectively, in equirectangular views. Thus, when dealing with such detections, as illustrated in Figure \ref{fig:fig5}, instead of directly projecting the whole box, this study first divided each of the boxes into two parts along the centre line, and then projected them separately, so that the detection outputs can be displayed and evaluated correctly.

\subsubsection{Merging the Reprojected Bounding Boxes}

In the final step, to merge bounding boxes of objects that cannot be fully displayed in a single sub-image, the boxes with at least one side coinciding with a boundary of a sub-image were extracted. According to Figure \ref{fig:fig6}, first, those boxes that are entirely within the 30-degree-wide overlapped areas were removed, as they must be included in another bounding box with smaller distortions. Then, the boxes of the same category are combined, which are tangential to consecutive boundaries. At the same time, the score of each new bounding box was calculated as the area-weighted average.




\subsection{Improving Multiple Object Tracking on Panoramic Cycling Videos}

\begin{figure}
    \centering
    \includegraphics[width=.90\textwidth]{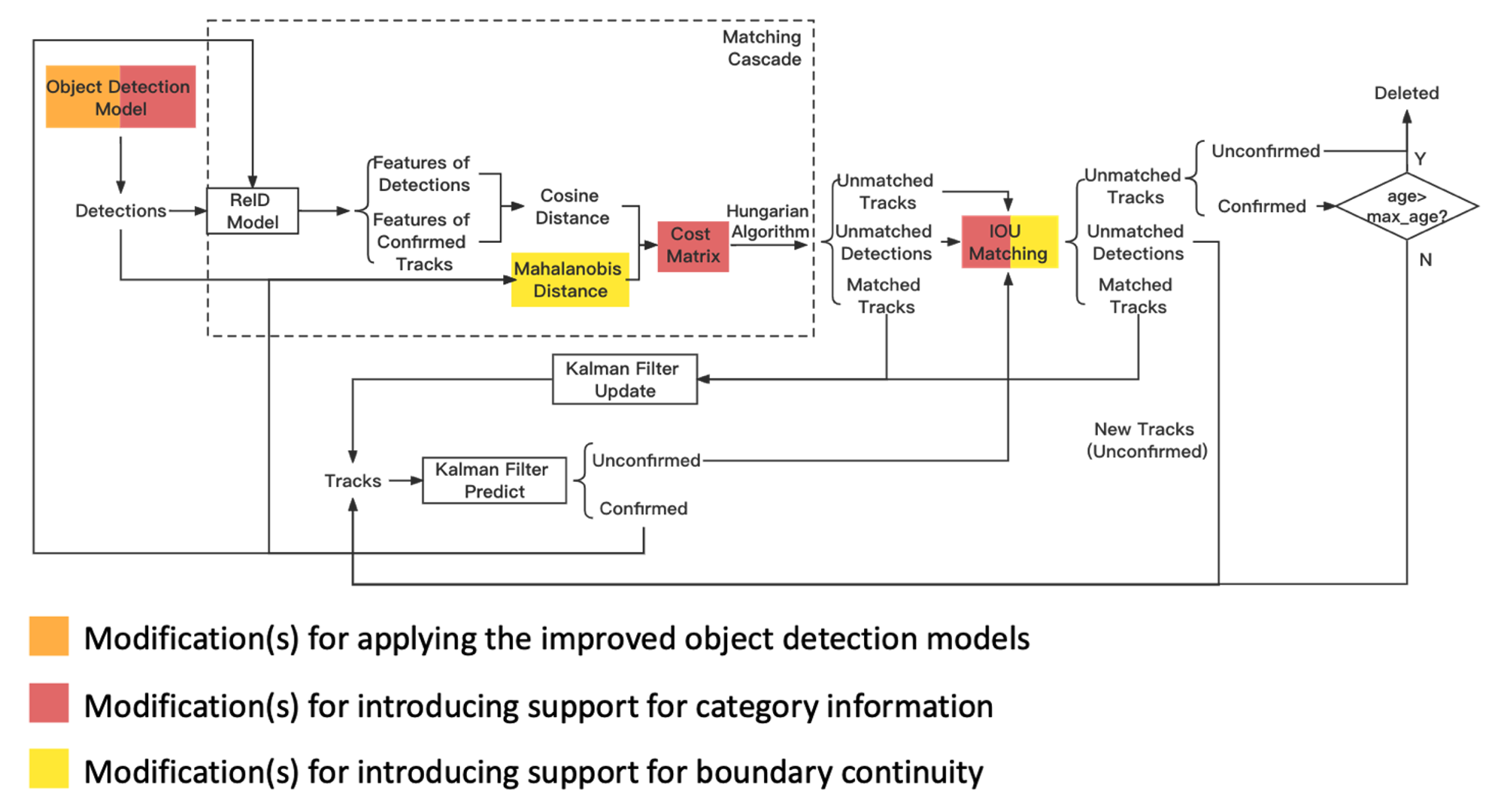}
    \caption{Workflow of improved object tracking model using StrongSORT as backbone for better tracking performance on panoramic cycling videos.}
    \label{fig:fig7}
\end{figure}

The second step of the proposed methodology is to improve the applicability of StrongSORT \citep{du2023strongsort}, one of the most recently proposed MOT models following the tracking-by-detection paradigm, to panoramic cycling videos. As shown in Figure \ref{fig:fig7}, the following modifications were done: First, improved Yolo12 object detection model was set as the detector; Then, support for category information and boundary continuity were introduced into the process of data association, according to the characteristics of panoramic videos. 

\subsubsection{StrongSORT}
According to \cite{du2023strongsort}, tracks in StrongSORT have two states: `confirmed' and `unconfirmed' --- only if a track gets matched in a certain number of consecutive frames will its state be switched to confirmed. For each frame in a video, StrongSORT sends $n$ detections and $m$ existing confirmed tracks into a vanilla global linear assignment module. The module extracts appearance features from each detection and updates the appearance state in an exponential moving average (EMA) manner proposed in \cite{wang2020towards}, instead of storing them with feature bank mechanism as in DeepSORT \citep{wojke2017simple}. The framework tracks with a pre-trained ReID model and calculates the appearance feature distance and Mahalanobis distance between them, whose weighted averages are then used to form an $m \times n$ cost matrix. Following this, the Hungarian algorithm is applied to associate the tracks and detection according to the values in the cost matrix, and the matched tracks are then updated by NSA Kalman filters \citep{du2021giaotracker} using the information of the corresponding detection. 

For the tracks and detections that are unmatched in the matching cascade module, StrongSORT creates another cost matrix according to their IOUs, and the Hungarian algorithm is used again to carry out data association. Next, StrongSORT creates new tracks for each unmatched detection, treating them as objects new to the video; while unmatched tracks are removed from the track list if they are `unconfirmed' or have not been matched for more than a certain number of frames. 

\subsubsection{Introducing the Support for Category Information}

As mentioned above, when associating the detections with tracks, StrongSORT ignores the category information of the detection, treating objects of different categories equally. Thus, tracks and detections can be wrongly matched when their corresponding values in the cost matrices are small. If this unmodified model is applied to panoramic cycling videos; such wrong associations can be seen when objects of different categories alternate at a short distance.

To address this issue, category information is included into the tracker. In this way, every track is initialized with the category label of the detection.
When computing cost matrices for matching, category filtering was performed. This involved setting values representing the distances between tracks and new detections with different or non-adjacent categories to an arbitrarily large number. This ensures that the existing tracks are only matched with the detections of the same or adjacent categories. Classes of objects like cars and trucks, bicycles and motorcycles are considered as adjacent categories. Sometimes the detector may assign an object its adjacent category in consecutive frames and then a track with incorrect category will be generated. By relaxing the condition of penalisation to adjacent categories, objects that are associated with an incorrect track can be re-matched with true category in later frames.
\subsubsection{Introducing the Support for Boundary Continuity}

\begin{figure}
    \centering
    \includegraphics[width=0.8\linewidth]{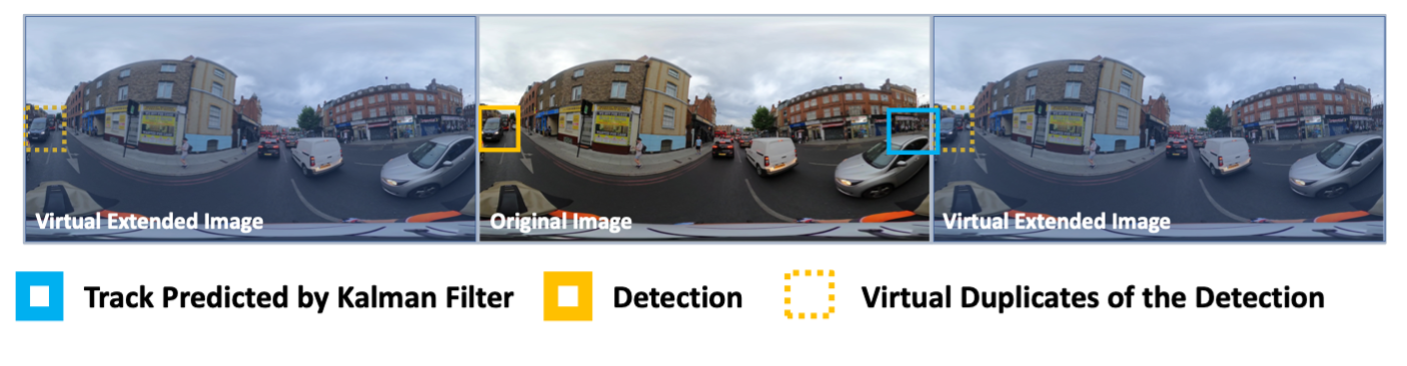}
    \caption{Preparations for the improved distance calculation process (suitable for Mahalanobis distance and IOU distance).}
    \label{fig:fig8}
\end{figure}

To track the detections which are divided into two parts (i.e., objects crossing the centre lines of the sub-images projected along 180$^\circ$), this research proposes an approach referring to the annotation method for panoramic MOT designed in subsection \ref{DataPreprocessing}: the first step is to move each of the left bounding boxes to the right and merge it with the right box; Then, input the combined bounding boxes into StrongSORT directly, despite the fact they are beyond the boundaries. 
In addition, to avoid ID switching caused by objects that leave from one side of the image and return from the other side, the distance calculation process of StrongSORT was also improved. As Figure \ref{fig:fig8} shows, for calculating the Mahalanobis or IOU distance between a predicted track and a detection, the proposed method first duplicates the detection on two virtual images extended from both sides of the original image; Then, it calculates the distances from the track to these three bounding boxes respectively and selects the shortest one to fill the cost matrix.

\subsection{Automated Overtaking Behaviour Detection}

\begin{figure}
    \centering
    \includegraphics[width=0.8\linewidth]{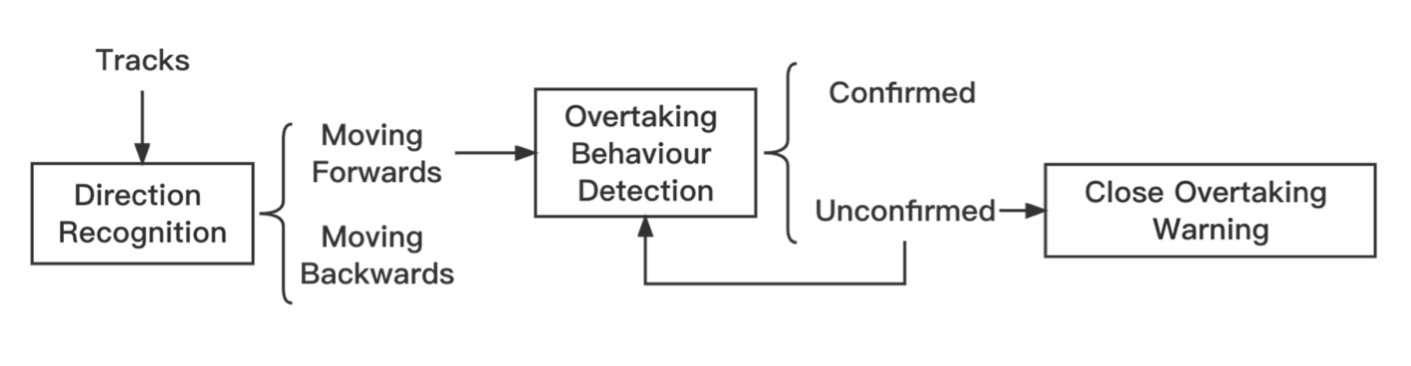}
    \caption{Workflow of the application which can detect the overtaking behaviour of the surrounding vehicles.}
    \label{fig:fig9}
\end{figure}

In the last step of the proposed methodology, an application was developed that uses the tracking results to detect a vehicle's overtaking behaviour.  As shown in Figure \ref{fig:fig9}, for each image frame, the application first classifies each vehicle into `forwards' or `backwards' according to the movement of its track. Following this, overtaking behaviour is detected among the vehicles moving forwards, and the detected overtakes are divided into two states: confirmed (completed) and unconfirmed (ongoing) --- For the confirmed overtakes, the application outputs their start and end frames with the track IDs; while for the unconfirmed ones, the application can warn the users of their current directions, so that the cyclists can react to the potential dangers in advance. It should be noted that overtaking is used as a test case to demonstrate the model, and different actions can be detected by changing the type, direction and location (left or right of the rider) of the tracked objects.


In the following subsections, each application module is described in detail and an experiment is carried out to evaluate the detection accuracy of the application.

\subsubsection{Movement Direction Recognition}

According to the definition of equirectangular projection \citep{snyder1997flattening}, objects behind the camera are shown on the left and right sides in a panoramic video. In contrast, objects in the front are shown in the middle. Therefore, if an object's current position is closer to the centre line than its position in the last frame, it is moving forward relative to the camera. Similarly, if an object moves away from the centre, it is moving backward.  

However, in practice, judging the moving directions based on only two frames is not rigorous: In some frames where the camera moves slightly with the rotation of the cyclists' head, some objects may incorrectly move in the opposite direction. Thus, to judge the movement of an object, this work calculated its movement in each of the last 5 frames relative to the current frame. If all of them were forward, it was classified as `forwards'. Likewise, if all of the movements were backwards, then the object was classified as `backwards'. 

\subsubsection{Overtaking Behaviour Detection}

The project defined overtaking behaviour as the act of a motor vehicle moving forwards and completely passing the cyclist. In an equirectangular video, this definition can be abstracted as the occasion in which the front and rear of a vehicle on the left (or right) side of the image cross the -90$^\circ$ (or 90$^\circ$) longitude line in order.

\begin{figure}[!ht]
    \centering
    \includegraphics[width=.80\textwidth]{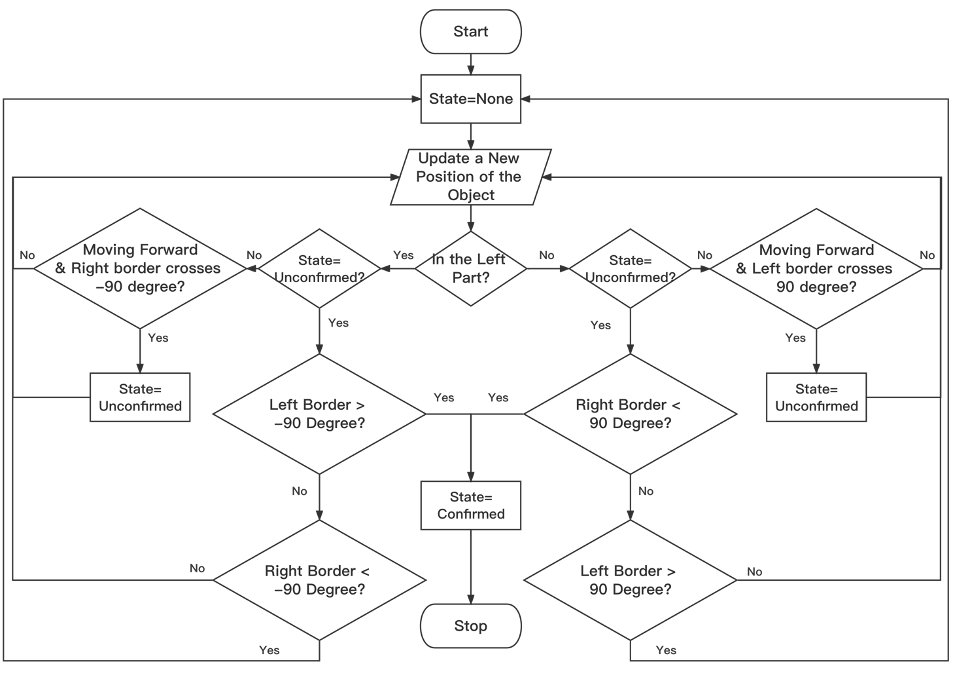}
    \caption{Flow chart of the overtaking behaviour detection for a specific object.}
    \label{fig:fig10}
\end{figure}

Figure \ref{fig:fig10} shows how the application was designed to realise the overtaking behaviour detection on a specific object in the panoramic videos: Take a vehicle on the left side of the video as an example, if the right side of its bounding box crosses the -90$^\circ$ longitude line for the first time as it is moving forwards, it will be assigned with an unconfirmed (ongoing) overtake. The overtake will remain unconfirmed until the left boundary of the bounding box passes the -90$^\circ$ line or the right border goes back to the left of the line, which means the overtake is finished (confirmed) or failed, respectively.

\section{Experiments and Results} \label{Results}
The experiments were conducted on a linux server equipped with two NVIDIA TITAN RTX GPUs. This chapter will present the details of experiments and analyse the results systematically. 

\subsection{Evaluation of the Model for Improving Object Detection on Panoramic Cycling Images}

The COCO-style annotated ground truth is used to assess the performance of the pre-trained YOLO12n before and after the method was applied to the dataset described in Section \ref{Data} under different input resolution settings (640, 1280, 1920). Mean average precision (mAPs) and average precision (APs) metrics of different sizes were recorded and compared, as higher APs indicate better object detection performance.
In addition, to assess the effect of merging the bounding boxes, which is an innovation compared to the method in \cite{yang2018object}'s study, the performance of models with and without combining the bounding boxes is also compared. 

According to Table \ref{tab6}, thanks to the smaller distortions of perspective projection, under the same input resolution settings, the proposed method can improve the detection performance for small size (0, $128^2$) objects. However, since multiple projections split some long objects into several parts, before applying the operation of merging the bounding boxes, the proposed models' precision metrics for medium and large size objects are lower than those of the original models with input resolution 640 and 1280. Merging the bounding boxes can effectively mitigate this problem for medium size objects, as APm increases after merging no matter what input resolution is used. However, the improved precision for large objects is still slightly smaller than that of the original models. 

Nevertheless, the majority of objects in the collected videos are small or medium sized, in most cases,  the mean APs of the proposed models exceed those of the original models, approaching the performance of the original models under higher input resolution settings.

\begin{table*}[h]
\caption{Average Precision Metrics of YOLO12n before and after the Proposed Methods are Applied. Bolded values represent the best results.\label{tab6}}
\centering
\small

\begin{tabular*}{\textwidth}{@{\extracolsep\fill}l c l r r r r r r@{}}\toprule

\textbf{\thead{Pre-Trained Model}} 
& \textbf{\thead{Input size}} 
& \textbf{Model} 
& \textbf{\(\text{AP}_{50:95}\)} 
& \textbf{\(\text{AP}_{50}\)} 
& \textbf{\(\text{AP}_{75}\)} 
& \textbf{APs**} 
& \textbf{APm**} 
& \textbf{APl**} 
\\
\midrule

\multirow{9}{*}{\makecell[l]{YOLO\\12n}} 
& \multirow{3}{*}{640} 
& Original 
& 13.2 & 25.9 & 11.5 & 6.6 & \textbf{30.8} & \textbf{35.5} \\
& 
& Proposed (No Merging) 
& 7.6 & 13.4 & 7.5 & 2.5 & 19.8 & 26.5 \\
& 
& Proposed (Merging) 
& \textbf{14.7} & \textbf{29.4} & \textbf{12.8} & \textbf{9.5} & 28.5 & 25.5 \\

\cmidrule{2-9}

& \multirow{3}{*}{1280} 
& Original 
& \textbf{16.6} & \textbf{31.2} & 15.2 & 10.1 & \textbf{36.1} & \textbf{35.2} \\
& 
& Proposed (No Merging) 
& 15.5 & 26.8 & 15.9 & 10.7 & 33.4 & 25.8 \\
& 
& Proposed (Merging) 
& \textbf{16.6} & 29.8 & \textbf{15.9} & \textbf{11.9} & 35.7 & 21.8 \\

\cmidrule{2-9}

& \multirow{3}{*}{1920} 
& Original 
& 13.3 & 21.9 & 13.9 & 8.5 & 29.2 & \textbf{25.2} \\
& 
& Proposed (No Merging) 
& 18.7 & 32.5 & 19.1 & 15.5 & 35.9 & 18.9 \\
& 
& Proposed (Merging) 
& \textbf{19.4} & \textbf{34.1} & \textbf{19.4} & \textbf{15.8} & \textbf{37.4} & 20.4 \\

\bottomrule
\end{tabular*}

\begin{tablenotes}
\item[] ** APs, APm and APl are the average precision metrics of the small $(0,128^2)$, medium $[128^2,384^2)$, and large $[384^2,+\infty)$ objects.
\end{tablenotes}

\end{table*}

\subsection{Evaluation of the Model for Improving Multiple Object Tracking on Panoramic Cycling Videos}
To evaluate the proposed method for improving MOT on panoramic cycling videos, StrongSORT is applied before and after the improved detection model was applied, and before and after the support for category information and boundary continuity were introduced. Each model is used to track the objects of interest in the three annotated videos. Evaluation metrics of the models were calculated by comparing their results with the ground truth.


As shown in Table \ref{tab6}, YOLO12n achieves high accuracy with moderate memory usage when the input size is 1280. Therefore, it was used as the detector of StrongSORT in the following analyses.  Using the improved YOLO as detectors, models before and after the implementations of support for category information and boundary continuity are compared, whose results are shown in Table \ref{tab8}. 

\begin{table*}[h]%
\caption{Evaluation Metrics of StrongSORT before and after the Supports for Category Information and Boundary Continuity are introduced.\label{tab8}}
\centering
\small
\begin{tabular*}{\textwidth}{@{\extracolsep\fill}lrrrrrrrrr@{}}\toprule
\textbf{Model} 
& \textbf{IDF1$\uparrow$} 
& \textbf{IDP$\uparrow$} 
& \textbf{IDR$\uparrow$}
& \textbf{FP$\downarrow$} 
& \textbf{FN$\downarrow$} 
& \textbf{IDs$\downarrow$} 
& \textbf{FM$\downarrow$} 
& \textbf{MOTA$\uparrow$} 
& \textbf{MOTP$\uparrow$} 
\\
\midrule
\makecell[l]{Original YOLO + StrongSORT} 
& \textbf{51.5}\% & 76.1\% & \textbf{38.9}\% 
& 818 & \textbf{12285} & 101 & \textbf{304} 
& \textbf{43.7}\% & \textbf{0.201}
\\
\makecell[l]{Improved YOLO + StrongSORT} 
& 45.4\% & 70.4\% & 33.6\% 
& 1030 & 13265 & 141 & 470 
& 38.5\% & 0.191
\\
\makecell[l]{Improved YOLO + StrongSORT w/ SB$^{\tnote{\bf a}}$} 
& 46.0\% & 72.5\% & 33.8\% 
& 766 & 13272 & 134 & 471 
& 39.6\% & 0.191
\\
\makecell[l]{Improved YOLO + StrongSORT w/ SC$^{\tnote{\bf b}}$} 
& 45.5\% & 70.5\% & 33.6\% 
& 1021 & 13266 & 142 & 470 
& 38.5\% & 0.191
\\
\makecell[l]{Improved YOLO + StrongSORT w/ SB and SC}  
& 50.0\% & \textbf{78.8}\% & 36.6\% 
& \textbf{544} & 13100 & \textbf{90} & 373 
& 41.5\% & 0.190
\\
\bottomrule
\end{tabular*}
\begin{tablenotes}
\item[$^{\rm a}$] Support for Boundary Continuity
\item[$^{\rm b}$] Support for Category Information
\end{tablenotes}
\end{table*}
As shown in the table \ref{tab8}, incorporating category information reduces false positives. Consequently, IDF1 improves marginally from 45.4\% to 45.5\%, with the number of ID switches and MOTA remaining nearly unchanged. Although the improving effect on metrics seems relatively slight, observing the output videos of the models with and without the support for category information shows that some incorrect data matching has been successfully avoided. On the other hand, the inclusion of boundary support can significantly reduce the number of false positives by almost 25.6 per cent. The IDF1 and MOTA are also be improved by 0.6\% and 0.9\%, respectively.


Combining the introduction of category information and support for boundary continuity further enhances the performance of StrongSORT. The IDF1 rises up to 50.0\% which is comparable to original tracking model with plain YOLO. The number of false positives and ID switches was reduced considerably, decreasing from 818 to 544 and from 101 to 90, respectively. As indicated by the reduction in ID switches, the tracking continuity of StrongSORT is improved.

\subsection{Evaluation of the Automated Overtaking Behaviour Detection}

In this experiment, we applied our improved YOLO12n $+$ StrongSORT framework to overtaking behaviour detection. Here, only cars, trucks and buses were selected as overtaking vehicles due to their more frequent involvement in crashes involving cyclists. The improved algorithm was tested on one hundred 15-second $360^{\circ}$ video clips collected under different riding conditions in London, UK, as mentioned in previous \autoref{Data}. The input resolution for YOLO12n is 1280. Table \ref{tabVal100} presents the test results as a confusion matrix.

\begin{table*}[h]%
\caption{Performance of overtaking behaviour detection using proposed detection framework powered by YOLO + StrongSORT.\label{tab:overtaking}}
\centering
\small
\begin{tabular*}{\textwidth}{@{\extracolsep\fill}lrrrrrrrr@{}}\toprule
\textbf{Underlying Tracking Model} 
& \textbf{Accuracy$\uparrow$} 
& \textbf{Precision$\uparrow$} 
& \textbf{Recall$\uparrow$} 
& \textbf{F1-score$\uparrow$} 
& \textbf{TP$\uparrow$}
& \textbf{TN$\uparrow$}
& \textbf{FP$\downarrow$}
& \textbf{FN$\downarrow$}
\\
\midrule
\makecell[l]{Improved YOLO + StrongSORT} 
& 79.0\% & 80.9\% & 76.0\% & 78.4\%
& 38 & 41 & 9 & 12
\\
\makecell[l]{Improved YOLO + StrongSORT w/ SB$^{\tnote{\bf a}}$} 
& 78.0\% & 79.2\% & 76.0\% & 77.6\%
& 38 & 40 & 10 & 12
\\
\makecell[l]{Improved YOLO + StrongSORT w/ SB and SC}  
& \textbf{82.0}\% & \textbf{83.3}\% & \textbf{80.0}\% & \textbf{81.6}\%
& \textbf{40} & \textbf{42} & \textbf{8} & \textbf{10}
\\
\bottomrule
\end{tabular*}
\begin{tablenotes}
\item[$^{\rm a}$] Support for Boundary Continuity
\item[$^{\rm b}$] Support for Category Information
\end{tablenotes}
\label{tabVal100}
\end{table*}

\begin{figure}[!htbp]          
  \centering
  \includegraphics[width=\textwidth]{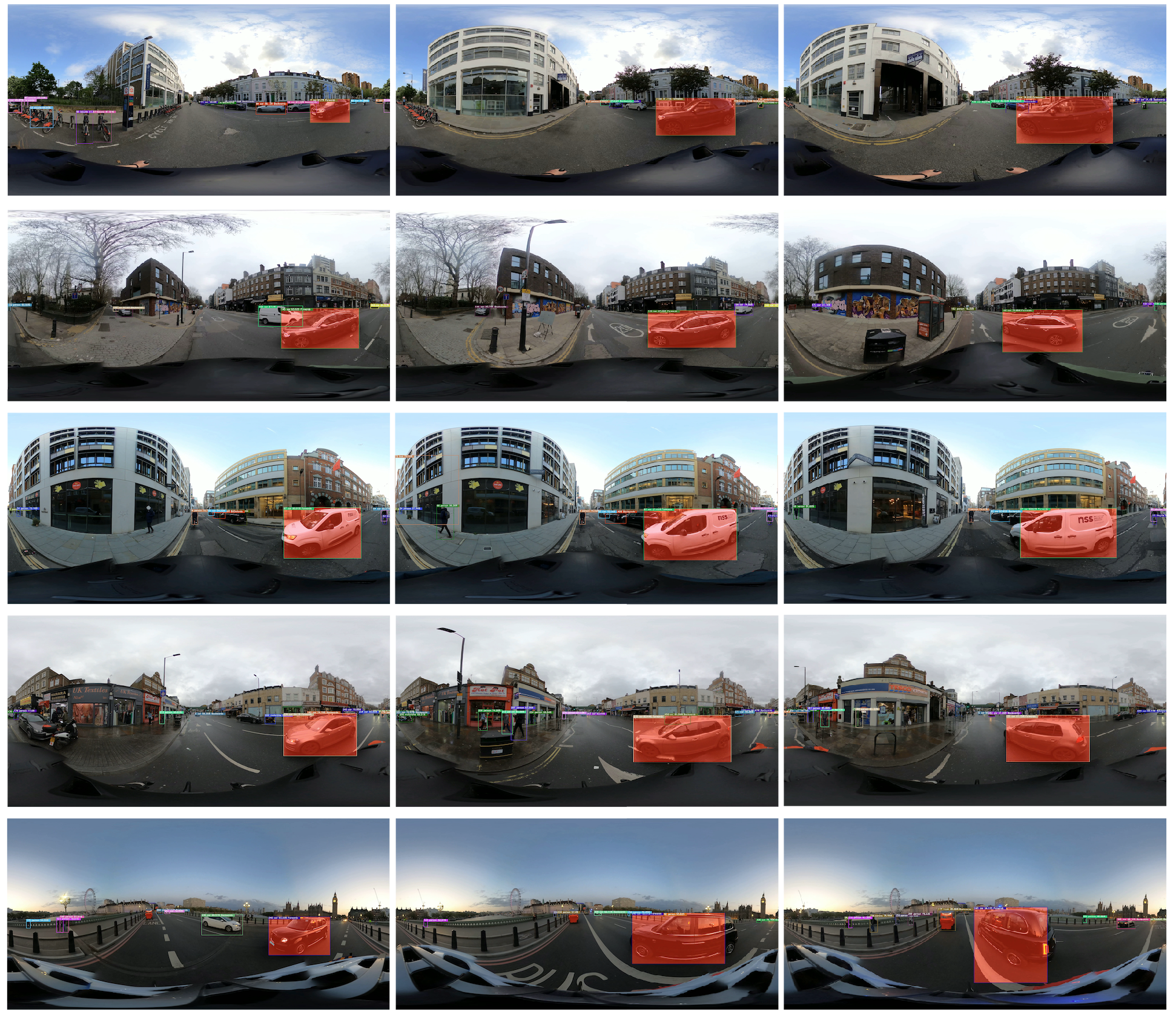}
  \caption{This figure illustrates five representative examples of automatically detected overtaking behaviours from the cyclist’s egocentric perspective. Each row corresponds to a single overtaking event and consists of three consecutive frames, in which the overtaking vehicle is highlighted by a red bounding box.}
  \label{fig:demo}
\end{figure}

It can be observed that Boundary Support and Category Support for the tracking model (in our case StrongSORT) achieves the best performance for the downstream overtaking detection task with a precision of 
\(
\text{Precision} = TP/(TP+FP) = 40/(40+8) = 0.83,
\)
and a recall (sensitivity) of 
\(
\text{Recall} = TP/(TP+FN) = 40/(40+10) = 0.80
\).
The F-score is computed as
\(
F\text{-score} = 2 \times \text{Precision} \times \text{Recall}/(\text{Precision} + \text{Recall}) = 2 \times 0.83 \times 0.80/(0.83+0.80) = 0.81
\). Removing SB and SC results in a modest but consistent decline in overtaking detection performance, with accuracy dropping from 82.0\% to 79.0\%, precision from 83.3\% to 80.9\%, recall from 80.0\% to 76.0\%, and F1-score from 81.6\% to 78.4\%. Meanwhile, as the qualitative study shown in Figure \ref{fig:demo}, the overtaking detection can demonstrate a decent performance on detecting positive overtaking events. 

\begin{figure}[!htbp]          
  \centering
  \includegraphics[width=\textwidth]{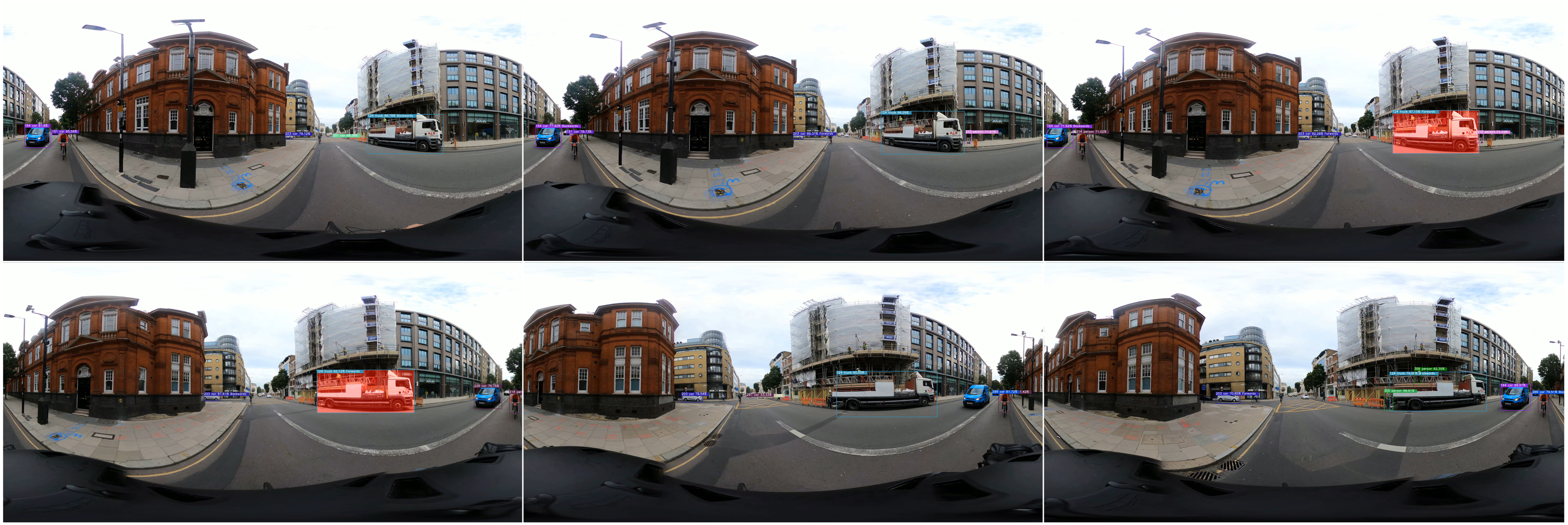}
  \caption{This figure contains six consecutive frames representing an falsely detected overtaking event of a truck. The cyclist was checking over the right shoulder, which made the truck appear moving towards the centre of the image. In consequence, the overtaking detector misunderstood the moving direction of the truck covered with red boxes in the third and fourth frame.}
  \label{fig:fig11}
\end{figure}

The 8 false positives are attributed to the cyclist checking over the shoulder, cyclist turning right (or left), and false positive detection of a car at night. The cyclist checking over the shoulder is the main cause of false positives for overtaking. 
For example in Figure \ref{fig:fig11}, a truck is moving backward relative to the bike but it appears to be moving forward and is determined as overtaking the cyclist when the cyclist checks over the shoulder to the right. Cyclists turning left or right can also cause a false positive for overtaking for a similar reason as checking over the shoulder. The camera's orientation will change when turning and hence static vehicles or those moving backward can be mistaken as moving forward and even overtaking. This issue can be addressed by fixing the orientation of the video relative to the trajectory of the rider and is the subject of ongoing work.

\begin{figure}[!htbp]          
  \centering
  \includegraphics[width=\textwidth]{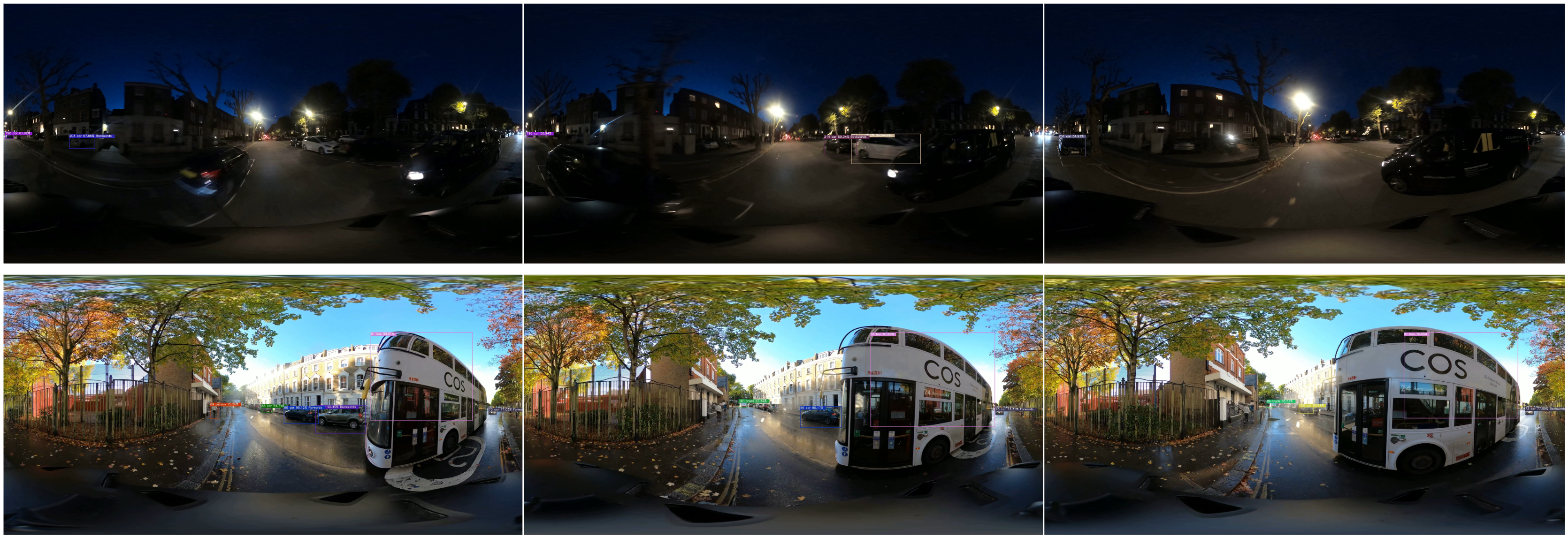}
  \caption{The three consecutive frames in the first row demonstrate an overtaking event in the night by a black car which is not detected at all. The three consecutive frames in the second row show an overtaking event by a bus which can be partly detected and tracked but its overtaking action is not detected.}
  \label{fig:fig12}
\end{figure}

Among the 10 videos classed as false negatives (where overtaking is undetected by the algorithm), in 9 videos the undetected overtaking vehicle is either a bus or black car moving during night-time. The detector cannot effectively distinguish black cars at night. In consequence, the model simply does not keep track of those black cars and cannot detect overtaking under this condition, as shown in the first row of Figure \ref{fig:fig12}. In cases involving undetected overtakes by buses, the bus can be detected and tracked. However, the boundaries of bounding boxes of buses are unstable, which sometimes invalidates the rule-based overtaking detection algorithm. An example is shown in the second row of Figure \ref{fig:fig12}, where the bus is detected but the bounding boxes are not accurate. Although the bus can be partly tracked with deviated bounding boxes, its relative moving direction can be misunderstood by the overtaking detection model.


\section{Discussion} \label{Discussion}

\subsection{Key Contributions}


Although Table \ref{tab6} indicates that the performance of the proposed method with input resolution 640 can also be achieved by the original models with larger input resolution, once the input resolution is as large as 1280 or 1920, the effect can not be replaced anymore. In other words, it increases the upper limit of accuracy that an object detection model can achieve on a panoramic image. Conversely, since the performance of the proposed object detection model on small and medium objects is always similar to (or even better than) that of the original models with higher input resolution, for the same accuracy requirements, the proposed model consumes fewer computing resources. Thus, it can achieve competitive accuracy even on low-end GPUs \citep{yang2018object}. 



The proposed tracking method for panoramic videos produces the annotation and results output of the objects which go across the boundaries in an equirectangular image without violating the MOT16 annotation rules \citep{milan2016mot16benchmarkmultiobjecttracking}. In this way, the existing MOT metrics can still be used for evaluating the tracking results on panoramic videos. Moreover, based on the performance improvement brought by the proposed object detection models, the introduction of category information and support for boundary continuity further enhances the applicability of StrongSORT to panoramic videos by reducing the number of false positives and ID switches. 

Taking advantage of the essence of equirectangular projection, which projects the front half of a spherical image to -90 degree to 90 degree and the rear half to the rest, the proposed overtaking detection framework is designed based on the location of bounding boxes of tracked objects, i.e., whether the tracked vehicle passes a certain angle threshold in the projected image. This rule-based algorithm can detect the completed overtaking behaviour of vehicles in panoramic videos with a high degree of accuracy and can output the start and end frames of each overtake with the corresponding track ID. This precise and automated analysis can provide abundant conflict data for studies on cycling safety in the future.


\subsection{Limitations and Future Work}

Despite its demonstrated performance, the proposed methodology has some limitations, which will be addressed in future work. Firstly, as multiple projection transformations divide long objects into several parts, according to Table \ref{tab6}, even when bounding box combination has been applied, the proposed models' performance on large objects (size $> 384^2$  pixels) is inferior to that of the original models. This can be explained by some parts of large objects (such as the blue car in sub-image 1) not being detected at all, preventing them from being merged. This issue may be mitigated by increasing the number of sub-images to explore whether more overlapping areas can achieve better detection performance on large objects. Furthermore, as predictions on sub-images are done sequentially, the inference speed is lower than the original models. This affects the operating speed of MOT and overtaking behaviour detection, which uses its results. Multi-processing can be used to address this in future work. This will allow inferences on sub-images in parallel. 



Similar to its predecessor DeepSORT, StrongSORT also uses a deep appearance ReID network trained on pedestrian datasets \citep{wojke2017simple, du2023strongsort}. Although \cite{zuraimi2021vehicle}'s research has demonstrated that it is possible to use a pretrained ReID descriptor on pedestrian images to extract appearance features on other objects such as vehicles, the results presented here show that when the angle of a vehicle with respect to the camera changes, some ID switching may occur. Referring to \cite{Li2021}, this research plans to construct a vehicle image dataset and re-train a ReID network with it, which may improve the performance of state-of-the-art tracking algorithms like StrongSORT in associating the vehicles according to their appearance.

The improved tracking model is validated on a downstream overtaking task, which relies on the information of the moving direction and position of other road users. Currently, the moving directions of the vehicles are estimated relative to the cyclists' movement. Since cyclists generally cannot move backward, it is reasonable to use relative forward movements to detect overtaking. However, for the `backward' objects (those moving from in front of to behind the rider), there are 3 possibilities for their actual movements: moving backward, being stationary or moving forward at a slower pace than the cyclist. Thus, if there is no other information such as the cyclist's speed, other road user behaviours such as braking or reversing may be challenging to detect. Fortunately, the GoPro Max camera, along with many consumer-grade action cameras, collects high-quality IMU and GPS data, including the user's speed, position, acceleration, and orientation. These data can be combined with the video to detect other behaviours associated with cycling risk, such as passing parked vehicles, filtering through traffic or entering the blind spot of heavy goods vehicles (HGVs). 

\section{Conclusion} \label{Conclusion}

In cycling safety, most existing studies applied computer vision models to `traditional' videos with a limited field of view for automated road user behaviour analysis. To enable analysis of the complete environment surrounding the rider, this research demonstrates an end-to-end approach for analysing cyclist safety by combining object detection and MOT in panoramic equirectangular videos. Existing pre-trained object detection models are applied to four perspective sub-images projected from equirectangular images. When evaluated on the annotated panoramic dataset, the performance of YOLO12n is improved under any input resolution setting. For MOT, a method that improved YOLO is used as the detector and support for boundary continuity and category information is implemented in StrongSORT, reducing false positives by 33.5\% and id switches by 10.9\%. The tracking results of the improved StrongSORT are implemented in the case of automated overtaking detection, which achieves high recall and precision on the testing videos.

Although the proposed methodology has been applied to overtaking detection, the framework can detect and track objects defined as classes in a pre-trained object detection model, such as pedestrians, bicycles, buses and motorcycles. Therefore, in the policy perspectives, the method can quantify compliance with minimum passing-distance policies and identify contexts (time of day, vehicle class) with elevated risk --- e.g., challenges detecting black cars at night and bounding-box instability for buses highlight specific night-time and large-vehicle contexts where targeted enforcement and operator training may be most beneficial. Moreover, fine-grained evidence on where passes initiate and complete relative to the rider can inform design details (lane widths, buffers, separation types), signal timings, and bus-stop bypasses. The boundary-aware tracking particularly helps follow vehicles across the full 360$^\circ$, supporting evidence on mid-block and junction approaches. With lightweight deployment, tracking outputs can back real-time rider alerts --- e.g., early warnings of a vehicle preparing to overtake or a “door-zone” advisory --- alongside post-ride feedback that helps cyclists adjust lane position or route choice. The paper outlines this real-time potential and the need for efficient models on edge devices .  

Overall, agencies and practitioners can use the proposed 360$^\circ$ detection-tracking-overtaking pipeline to generate near-miss surrogates from routine rides, map and monitor hazardous passing, evaluate infrastructure and policy changes when crashes are rare, and support both targeted enforcement and rider-facing warning systems --- accelerating evidence-led cycling safety decisions.  However, two technical issues need to be highlighted as the current limitations: (i) detection of large objects occasionally degrades because projections split them across sub-images. Increasing overlap or sub-image count may help, balanced against runtime, and (ii) night-time/black vehicles and buses can challenge the overtaking rule set due to missed detections and unstable boxes.







\bmsection*{Data availability statement}
The video data from the 100 Cyclists Project used in this study are not publicly available to protect the privacy of the participants. A sample video collected by the research team is available on the project's Github page at \href{https://github.com/SpaceTimeLab/360_object_tracking}{https://github.com/SpaceTimeLab/360\_object\_tracking}, along with the code used to generate the results.

\bmsection*{Acknowledgments}
The authors would like to thank the participants of the 100 Cyclists Project for volunteering their time and contributing valuable data to this research. They would also like to thank the reviewers for their feedback, which have helped to improve the manuscript.



\bmsection*{Conflict of interest}

The authors declare no potential conflict of interests.

\bibliography{wileyNJD-Harvard}

@article{gao2025reliable,
	title        = {Reliable imputation of incomplete crash data for predicting driver injury severity},
	author       = {Gao, Xiaowei and Jiang, Xinke and Zhuang, Dingyi and Haworth, James and Wang, Shenhao and Ilyankou, Ilya and Chen, Huanfa},
	year         = 2025,
	journal      = {Accident Analysis \& Prevention},
	publisher    = {Elsevier},
	volume       = 216,
	pages        = 108020
}

@book{brunsing_1997,
	title        = {PUBLIC TRANSPORT AND CYCLING: EXPERIENCE OF MODAL INTEGRATION IN GERMANY. FROM THE GREENING OF URBAN TRANSPORT},
	author       = {Brunsing, J.},
	year         = 1997,
	publisher    = {Wiley (John) and Sons, Limited},
	edition      = 2
}

@article{Daniels2008,
	title        = {The effects of roundabouts on traffic safety for bicyclists: An observational study},
	author       = {Daniels,  Stijn and Nuyts,  Erik and Wets,  Geert},
	year         = 2008,
	journal      = {Accident Analysis \& Prevention},
	publisher    = {Elsevier BV},
	volume       = 40,
	number       = 2,
	pages        = {518–526},
	issn         = {0001-4575}
}

@article{Delmelle2012,
	title        = {Exploring spatio-temporal commuting patterns in a university environment},
	author       = {Delmelle,  Eric M. and Delmelle,  Elizabeth Cahill},
	year         = 2012,
	journal      = {Transport Policy},
	publisher    = {Elsevier BV},
	volume       = 21,
	pages        = {1–9},
	issn         = {0967-070X}
}

@inproceedings{deng_2017,
	title        = {Object detection on panoramic images based on deep learning},
	author       = {Fucheng Deng and Xiaorui Zhu and Jiamin Ren},
	year         = 2017,
	booktitle    = {2017 3rd International Conference on Control, Automation and Robotics (ICCAR)},
	volume       = {},
	number       = {},
	pages        = {375--380},
	doi          = {10.1109/ICCAR.2017.7942721},
	organization = {IEEE}
}

@article{Hels2007,
	title        = {The effect of roundabout design features on cyclist accident rate},
	author       = {Hels,  Tove and Orozova-Bekkevold,  Ivanka},
	year         = 2007,
	journal      = {Accident Analysis \& Prevention},
	publisher    = {Elsevier BV},
	volume       = 39,
	number       = 2,
	pages        = {300–307},
	issn         = {0001-4575}
}

@inproceedings{hong_hua_2001,
	title        = {A high-resolution panoramic camera},
	author       = {Hong Hua and N. Ahuja},
	year         = 2001,
	booktitle    = {Proceedings of the IEEE Conference on Computer Vision and Pattern Recognition (CVPR)},
	volume       = 1,
	pages        = {I--I},
	doi          = {10.1109/CVPR.2001.990634},
	organization = {Proceedings of the IEEE Conference on Computer Vision and Pattern Recognition (CVPR)}
}

@article{Ibrahim2021,
	title        = {Cycling near misses: a review of the current methods,  challenges and the potential of an AI-embedded system},
	author       = {Ibrahim,  Mohamed R. and Haworth, James and Christie, Nicola and Cheng, Tao and Hailes, Stephen},
	year         = 2021,
	journal      = {Transport Reviews},
	publisher    = {Informa UK Limited},
	volume       = 41,
	number       = 3,
	pages        = {304–328},
	issn         = {1464-5327}
}

@inproceedings{Im2016,
  title={All-around depth from small motion with a spherical panoramic camera},
  author={Im, Sunghoon and Ha, Hyowon and Rameau, Fran{\c{c}}ois and Jeon, Hae-Gon and Choe, Gyeongmin and Kweon, In So},
  booktitle={European Conference on Computer Vision},
  pages={156--172},
  year={2016},
  organization={Springer}
}

@article{JimnezBravo2022,
	title        = {Multi-object tracking in traffic environments: A systematic literature review},
	author       = {Jiménez-Bravo,  Diego M. and Lozano Murciego,  Alvaro and Sales Mendes,  André and Sánchez San Blás,  Héctor and Bajo,  Javier},
	year         = 2022,
	journal      = {Neurocomputing},
	publisher    = {Elsevier BV},
	volume       = 494,
	pages        = {43–55},
	issn         = {0925-2312}
}

@inproceedings{jodoin_2014,
	title        = {Urban Tracker: Multiple object tracking in urban mixed traffic},
	author       = {Jodoin, Jean-Philippe and Bilodeau, Guillaume-Alexandre and Saunier, Nicolas},
	year         = 2014,
	booktitle    = {IEEE Winter Conference on Applications of Computer Vision},
	volume       = {},
	number       = {},
	pages        = {885--892},
	doi          = {10.1109/WACV.2014.6836010},
	organization = {IEEE}
}

@article{johnson_2010,
	title        = {Naturalistic cycling study: identifying risk factors for on-road commuter cyclists},
	author       = {Marilyn Johnson and Judith Charlton and Jennifer Oxley and Stuart Newstead},
	year         = 2010,
	journal      = {Annals of Advances in Automotive Medicine},
	volume       = 54,
	pages        = {275--283},
	issn         = {1943-2461},
	note         = {Annual Conference of the Association for the Advancement of Automotive Medicine (AAAM) 2010; Conference date: 01-01-2010},
	language     = {English}
}

@article{LeCun2015,
	title        = {Deep learning},
	author       = {LeCun,  Yann and Bengio,  Yoshua and Hinton,  Geoffrey},
	year         = 2015,
	journal      = {Nature},
	publisher    = {Springer Science and Business Media LLC},
	volume       = 521,
	number       = 7553,
	pages        = {436–444},
	issn         = {1476-4687}
}

@article{Lee2021,
	title        = {Viewport Rendering Algorithm with a Curved Surface for a Wide FOV in 360$^\circ$ Images},
	author       = {Lee,  Geon-Won and Han,  Jong-Ki},
	year         = 2021,
	journal      = {Applied Sciences},
	publisher    = {MDPI AG},
	volume       = 11,
	number       = 3,
	pages        = 1133,
	issn         = {2076-3417}
}

@inproceedings{tian2025yolov12,
	title        = {YOLOv12: Attention-Centric Real-Time Object Detectors},
	author       = {Tian, Yunjie and Ye, Qixiang and Doermann, David},
	year         = 2025,
	booktitle    = {The Thirty-ninth Annual Conference on Neural Information Processing Systems},
	organization = {The Thirty-ninth Annual Conference on Neural Information Processing Systems}
}

@inproceedings{Li2021,
	title        = {Research on Distribution Strategy of Region of Interest in Panoramic Video Based on Improved Deepsort},
	author       = {Li, Jiangeng and Zang, Zhibo and Xie, Haizheng and Wang, Guangsheng},
	year         = 2021,
	booktitle    = {2021 33rd Chinese Control and Decision Conference (CCDC)},
	pages        = {1714--1719},
	organization = {IEEE}
}

@inproceedings{lin_2014,
	title        = {Microsoft COCO: Common Objects in Context},
	author       = {Lin, Tsung-Yi and Maire, Michael an Belongie, Serge and Hays, James and Perona, Pietro and Ramanan, Deva and Dollar, Piotr and Zitnick, C. Lawrence},
	year         = 2014,
	booktitle    = {European conference on computer vision},
	organization    = {Cham: Springer International Publishing},
	pages        = {740--755}}

@inproceedings{liu_2018,
	title        = {Simple online and realtime tracking with spherical panoramic camera},
	author       = {Liu, Keng-Chi and Shen, Yi-Ting and Chen, Liang-Gee},
	year         = 2018,
	booktitle    = {2018 IEEE International Conference on Consumer Electronics (ICCE)},
	volume       = {},
	number       = {},
	pages        = {1--6},
	doi          = {10.1109/ICCE.2018.8326132},
	organization = {IEEE}
}

@article{Luo2021,
	title        = {Multiple object tracking: A literature review},
	author       = {Luo,  Wenhan and Xing,  Junliang and Milan,  Anton and Zhang,  Xiaoqin and Liu,  Wei and Kim,  Tae-Kyun},
	year         = 2021,
	journal      = {Artificial Intelligence},
	publisher    = {Elsevier BV},
	volume       = 293,
	pages        = 103448,
	doi          = {10.1016/j.artint.2020.103448},
	issn         = {0004-3702}
}

@inproceedings{masalov_2018,
	title        = {CyDet: Improving Camera-based Cyclist Recognition Accuracy with Known Cycling Jersey Patterns},
	author       = {Masalov, Alexerand and Ota, Jeffrey and Corbet, Heath and Lee, Eric and Pelley, Adam},
	year         = 2018,
	booktitle    = {2018 IEEE Intelligent Vehicles Symposium (IV)},
	location     = {Changshu, Suzhou, China},
	publisher    = {IEEE Press},
	pages        = {2143–2149},
	doi          = {10.1109/IVS.2018.8500668},
	organization = {IEEE},
	numpages     = 7
}

@article{massink_2011,
	title        = {The Climate Value of Cycling},
	author       = {Massink, Roel and Zuidgeest, Mark and Rijnsburger, Jaap and Sarmiento, Olga L. and van Maarseveen, Martin},
	year         = 2011,
	journal      = {Natural Resources Forum},
	volume       = 35,
	number       = 2,
	pages        = {100--111},
	doi          = {https://doi.org/10.1111/j.1477-8947.2011.01345.x},
	eprint       = {https://onlinelibrary.wiley.com/doi/pdf/10.1111/j.1477-8947.2011.01345.x}
}

@misc{milan2016mot16benchmarkmultiobjecttracking,
	title        = {MOT16: A Benchmark for Multi-Object Tracking},
	author       = {Anton Milan and Laura Leal-Taixe and Ian Reid and Stefan Roth and Konrad Schindler},
	year         = 2016,
	eprint       = {1603.00831},
	archiveprefix = {arXiv},
	primaryclass = {cs.CV},
	howpublished = {arXiv preprint}
}

@article{ojala2022motion,
	title        = {Motion detection and classification: ultra-fast road user detection},
	author       = {Ojala, Risto and Vepsalainen, Jari and Tammi, Kari},
	year         = 2022,
	journal      = {Journal of Big Data},
	publisher    = {Springer},
	volume       = 9,
	number       = 1,
	pages        = 28
}

@article{poulos2012exposure,
	title        = {Exposure-based cycling crash, near miss and injury rates: The Safer Cycling Prospective Cohort Study protocol},
	author       = {Poulos, Roslyn G and Hatfield, Julie and Rissel, Chris and Grzebieta, Raphael and McIntosh, Andrew S},
	year         = 2012,
	journal      = {Injury prevention},
	publisher    = {BMJ Publishing Group Ltd},
	volume       = 18,
	number       = 1,
	pages        = {e1--e1}
}

@article{pucher2016safer,
	title        = {Safer cycling through improved infrastructure},
	author       = {Pucher, John and Buehler, Ralph},
	year         = 2016,
	journal      = {American Journal of Public Health},
	publisher    = {American Public Health Association},
	volume       = 106,
	number       = 12,
	pages        = {2089--2091}
}

@book{snyder1997flattening,
  title={Flattening the earth: two thousand years of map projections},
  author={Snyder, John P},
  year={1997},
  publisher={University of Chicago Press}
}

@inproceedings{Taha2015,
	title        = {Multi Objects Tracking in Nighttime Traffic Scenes},
	author       = {Taha,  Mohamed and Zayed,  Hala H. and Nazmy,  Taymoor and Khalifa,  M. E.},
	year         = 2015,
	booktitle    = {The 7th International Conference on Information Technology},
	publisher    = {Al-Zaytoonah University of Jordan},
	series       = {ICIT 2015},
	doi          = {10.15849/icit.2015.0002},
	organization = {Al-Zaytoonah University of Jordan},
	collection   = {ICIT 2015}
}

@article{tait2022cycling,
	title        = {Is cycling infrastructure in London safe and equitable? Evidence from the cycling infrastructure database},
	author       = {Tait, Caroline and Beecham, Roger and Lovelace, Robin and Barber, Stuart},
	year         = 2022,
	journal      = {Journal of Transport \& Health},
	publisher    = {Elsevier},
	volume       = 26,
	pages        = 101369
}

@article{du2023strongsort,
	title        = {Strongsort: Make deepsort great again},
	author       = {Du, Yunhao and Zhao, Zhicheng and Song, Yang and Zhao, Yanyun and Su, Fei and Gong, Tao and Meng, Hongying},
	year         = 2023,
	journal      = {IEEE Transactions on Multimedia},
	publisher    = {IEEE},
	volume       = 25,
	pages        = {8725--8737}
}

@article{wanner2012active,
	title        = {Active transport, physical activity, and body weight in adults: a systematic review},
	author       = {Wanner, Miriam and G{\"o}tschi, Thomas and Martin-Diener, Eva and Kahlmeier, Sonja and Martin, Brian W},
	year         = 2012,
	journal      = {American journal of preventive medicine},
	publisher    = {Elsevier},
	volume       = 42,
	number       = 5,
	pages        = {493--502}
}

@inproceedings{wojke2017simple,
	title        = {Simple online and realtime tracking with a deep association metric},
	author       = {Wojke, Nicolai and Bewley, Alex and Paulus, Dietrich},
	year         = 2017,
	booktitle    = {2017 IEEE international conference on image processing (ICIP)},
	pages        = {3645--3649},
	organization = {2017 IEEE international conference on image processing (ICIP)}
}

@inproceedings{yang2018object,
	title        = {Object detection in equirectangular panorama},
	author       = {Yang, Wenyan and Qian, Yanlin and K{\"a}m{\"a}r{\"a}inen, Joni-Kristian and Cricri, Francesco and Fan, Lixin},
	year         = 2018,
	booktitle    = {2018 24th international conference on pattern recognition (icpr)},
	pages        = {2190--2195},
	organization = {IEEE}
}

@inproceedings{zhang2017real,
	title        = {Real-Time object detection for 360-degree panoramic image using CNN},
	author       = {Zhang, Yiming and Xiao, Xiangyun and Yang, Xubo},
	year         = 2017,
	booktitle    = {2017 international conference on virtual reality and visualization (icvrv)},
	pages        = {18--23},
	organization = {IEEE}
}

@article{zhao2019object,
	title        = {Object detection with deep learning: A review},
	author       = {Zhao, Zhong-Qiu and Zheng, Peng and Xu, Shou-tao and Wu, Xindong},
	year         = 2019,
	journal      = {IEEE transactions on neural networks and learning systems},
	publisher    = {IEEE},
	volume       = 30,
	number       = 11,
	pages        = {3212--3232}
}

@inproceedings{zuraimi2021vehicle,
	title        = {Vehicle detection and tracking using YOLO and DeepSORT},
	author       = {Zuraimi, Muhammad Azhad Bin and Zaman, Fadhlan Hafizhelmi Kamaru},
	year         = 2021,
	booktitle    = {2021 IEEE 11th IEEE Symposium on Computer Applications \& Industrial Electronics (ISCAIE)},
	pages        = {23--29},
    }

@article{BECK2019253,
	title        = {How much space do drivers provide when passing cyclists? Understanding the impact of motor vehicle and infrastructure characteristics on passing distance},
	author       = {Ben Beck and Derek Chong and Jake Olivier and Monica Perkins and Anthony Tsay and Adam Rushford and Lingxiao Li and Peter Cameron and Richard Fry and Marilyn Johnson},
	year         = 2019,
	journal      = {Accident Analysis \& Prevention},
	volume       = 128,
	pages        = {253--260},
	issn         = {0001-4575}
}

@inproceedings{du2021giaotracker,
	title        = {Giaotracker: A comprehensive framework for mcmot with global information and optimizing strategies in visdrone 2021},
	author       = {Du, Yunhao and Wan, Junfeng and Zhao, Yanyun and Zhang, Binyu and Tong, Zhihang and Dong, Junhao},
	year         = 2021,
	booktitle    = {Proceedings of the IEEE/CVF International conference on computer vision},
	pages        = {2809--2819}
}

@inproceedings{wang2020towards,
	title        = {Towards real-time multi-object tracking},
	author       = {Wang, Zhongdao and Zheng, Liang and Liu, Yixuan and Li, Yali and Wang, Shengjin},
	year         = 2020,
	booktitle    = {European conference on computer vision},
	pages        = {107--122},
	organization = {Springer}
}

@article{rocky_review_2024,
  title={Review of accident detection methods using dashcam videos for autonomous driving vehicles},
  author={Rocky, Arash and Wu, Qingming Jonathan and Zhang, Wandong},
  journal={IEEE Transactions on Intelligent Transportation Systems},
  volume={25},
  number={8},
  pages={8356--8374},
  year={2024},
  publisher={IEEE}
}

@article{rick_cycling_2021,
	title        = {Cycling in the {Flattened} {City}: {Urban} {Assemblages} and {Digital} {Visual} {Research}},
	shorttitle   = {Cycling in the {Flattened} {City}},
	author       = {Rick, Oliver J. C. and Bustad, Jacob J.},
	year         = 2021,
	journal      = {Somatechnics},
	volume       = 11,
	number       = 2,
	pages        = {246--264},
	issn         = {2044-0138},
	note         = {Publisher: Edinburgh University Press}
}

@article{wang2020exploring,
	title        = {Exploring factors influencing the risky cycling behaviors of young cyclists aged 15--24 years: A questionnaire-based study in China},
	author       = {Wang, Cheng and Zhang, Weihua and Feng, Zhongxiang and Wang, Kun and Gao, Yuhua},
	year         = 2020,
	journal      = {Risk analysis},
	publisher    = {Wiley Online Library},
	volume       = 40,
	number       = 8,
	pages        = {1554--1570}
}

@article{useche2022cross,
	title        = {Cross-culturally approaching the cycling behaviour questionnaire (CBQ): evidence from 19 countries},
	author       = {Useche, Sergio A and Alonso, Francisco and Boyko, Aleksey and Buyvol, Polina and Castaneda, Isaac and Cendales, Boris and Cervantes, Arturo and Echiburu, Tomas and Faus, Mireia and Feitosa, Zuleide and others},
	year         = 2022,
	journal      = {Transportation research part F: traffic psychology and behaviour},
	publisher    = {Elsevier},
	volume       = 91,
	pages        = {386--400}
}

@article{fischer2020does,
	title        = {What does crowdsourced data tell us about bicycling injury? A case study in a mid-sized Canadian city},
	author       = {Fischer, Jaimy and Nelson, Trisalyn and Laberee, Karen and Winters, Meghan},
	year         = 2020,
	journal      = {Accident Analysis \& Prevention},
	publisher    = {Elsevier},
	volume       = 145,
	pages        = 105695
}

@article{zhang2023analyzing,
	title        = {Analyzing the injury severity in single-bicycle crashes: An application of the ordered forest with some practical guidance},
	author       = {Zhang, Yingheng and Li, Haojie and Ren, Gang},
	year         = 2023,
	journal      = {Accident Analysis \& Prevention},
	publisher    = {Elsevier},
	volume       = 189,
	pages        = 107126
}

@article{ibrahim_cyclingnet_2021,
	title        = {{CyclingNet}: {Detecting} cycling near misses from video streams in complex urban scenes with deep learning},
	author       = {Ibrahim, Mohamed R. and Haworth, James and Christie, Nicola and Cheng, Tao},
	year         = 2021,
	journal      = {IET Intelligent Transport Systems},
	volume       = 15,
	number       = 10,
	pages        = {1331--1344},
	issn         = {1751-9578},
	copyright    = {© 2021 The Authors. IET Intelligent Transport Systems published by John Wiley \& Sons Ltd on behalf of The Institution of Engineering and Technology},
	language     = {en}
}

@article{Ambros2018,
	title        = {An international review of challenges and opportunities in development and use of crash prediction models},
	author       = {Ambros,  Jiří and Jurewicz,  Chris and Turner,  Shane and Kieć,  Mariusz},
	year         = 2018,
	journal      = {European Transport Research Review},
	publisher    = {Springer Science and Business Media LLC},
	volume       = 10,
	number       = 2,
	issn         = {1866-8887}
}

@techreport{australian_transport_safety_bureau_deaths_2006,
	title        = {Deaths of cyclists due to road crashes},
	author       = {{Australian Transport Safety Bureau}},
	year         = 2006,
	address      = {Canberra, Australia},
	urldate      = {2024-07-26},
	institution  = {Australian Transport Safety Bureau}
}

@techreport{transport_for_london_casualties_2020,
	title        = {Casualties in {Greater} {London} 2020},
	author       = {{Transport for London}},
	year         = 2020,
	address      = {London, UK},
	urldate      = {2024-07-26},
	language     = {en},
	institution  = {Transport for London}
}

@article{christiedetecting,
	title        = {Spatiotemporal and behavioural correlates of cycling near misses: evidence from helmet-mounted video data},
	author       = {Nicola Christie and James Haworth and Mohamed Ibrahim and Xiaowei Gao and Natchapon Jongwiriyanurak and Meihui Wang and Jingwei Guo},
	year         = 2026,
	journal      = {Accident Analysis \& Prevention},
	volume       = 224,
	pages        = 108272
}



\end{document}